\documentclass[letterpaper, 10 pt, conference]{ieeeconf}  % Comment this line out if you need a4paper

\IEEEoverridecommandlockouts                              % This command is only needed if 
\usepackage{graphics} % for pdf, bitmapped graphics files
\usepackage{epsfig} % for postscript graphics files
\usepackage{mathptmx} % assumes new font selection scheme installed
\usepackage{times} % assumes new font selection scheme installed
\usepackage{amsmath} % assumes amsmath package installed
\usepackage{amssymb}  % assumes amsmath package installed
\usepackage{hyperref}
\usepackage{xcolor}
\usepackage{tikz}
\usepackage[lofdepth,lotdepth]{subfig}

\usepackage{cite}
\usepackage{hyperref}
\hypersetup{
    colorlinks=true,
    linkcolor=blue,
    filecolor=magenta,      
    urlcolor=red,
    pdftitle={Overleaf Example},
    pdfpagemode=FullScreen,
    }

\usepackage[keeplastbox]{flushend} %to balance columns on last page

\usepackage[normalem]{ulem} %strike through

\usepackage{multirow}

\title{\LARGE \bf
Semantic State Estimation in  Cloth Manipulation Tasks
}

\author{Georgies Tzelepis$^{1}$,  Eren Erdal Aksoy$^{2}$, Júlia Borràs$^{1}$,  and Guillem Alenyà$^{1}$  % <-this % stops a space
 \thanks{*This work receives funding from the Spanish State Research Agency through the project CHLOE-GRAPH (PID2020-118649RB-l00); and the EU H2020 Programme under grant agreement ERC–2016–ADG–741930 (CLOTHILDE).}% <-this % stops a space
\thanks{$^{1}$ The authors are  with Institut de Robòtica i Informàtica Industrial, CSIC-UPC,  
		Llorens i Artigas 4-6, 08028 Barcelona, Spain. {\tt \{gtzelepis,jborras, galenya\}@iri.upc.edu}}%
\thanks{$^{2}$Halmstad University, Center for Applied Intelligent Systems Research, Halmstad, Sweden
        {\tt\small eren.aksoy@hh.se}}%
}

\begin{document}

\maketitle
\thispagestyle{empty}
\pagestyle{empty}

%%%%%%%%%%%%%%%%%%%%%%%%%%%%%%%%%%%%%%%%%%%%%%%%%%%%%%%%%%%%%%%%%%%%%%%%%%%%%%%%
\begin{abstract}

Understanding  of deformable object manipulations such as textiles is a challenge due to the complexity and high dimensionality of the problem.
Particularly, the lack of a generic representation of semantic states (e.g., \textit{crumpled}, \textit{diagonally folded}) during a continuous manipulation process introduces an obstacle to identify the manipulation type.
In this paper, we aim to solve the problem of semantic state estimation in cloth manipulation tasks.
For this purpose, we introduce a new large-scale fully-annotated RGB image dataset showing various human demonstrations of different complicated cloth manipulations. 
We provide a set of baseline deep networks and benchmark them on the problem of semantic state estimation using our proposed dataset.
Furthermore, we investigate the scalability of our semantic state estimation framework in robot monitoring tasks of long and complex cloth manipulations. 
We also release the dataset and source code of the baseline models:
\url{https://github.com/BOBaraki/state-estimation-for-continuous-cloth-manipulation}
% \textcolor{red}{\href{Insert project link}}.

\end{abstract}

%%%%%%%%%%%%%%%%%%%%%%%%%%%%%%%%%%%%%%%%%%%%%%%%%%%%%%%%%%%%%%%%%%%%%%%%%%%%%%%%
\section{INTRODUCTION}

Textiles are an important part of our daily living objects both in domestic, public health, and industrial scenarios.
While rigid object manipulation has achieved maturity, cloth manipulation remains in its infancy due to its high complexity, and only recently it is becoming a very active research topic. Recent results include novel manipulation solutions, mainly focused on cloth state estimation, grasp point selection, and efficient representations \cite{ doumanoglou2016folding,hoque2020visuospatial,lippi2020latent, pumarola2018geometry, corona2018active, ramisa20163d,seita2018deep,seita2020deep,matas2018sim}, but the high-level understanding of cloth deformation state is still an uncharted challenge. 
Unlike their rigid and articulated counterparts, where the number of possible states for an object is manageable and can be semantically defined and linked to actions, identifying semantic deformation states of a textile object is a high dimensional problem that has so far been unexplored, to the best of our knowledge. In robotics, recognizing the semantic state of a textile in a continuous manipulation is essential for the subsequent tasks, such as monitoring,  task planning, learning from human demonstrations, and action execution. %decision making.

%this paragraph is good, but shouldn't be here as it breaks the flow about semantic states.
% In computer vision, deep learning has become the primary approach for solving perception-related tasks in an end-to-end fashion. However, their superior performance requires large scale annotated datasets. 
% %
% This is also true in  the context of cloth manipulation as shown in works like \cite{tanaka2018emd,seita2020deep,jangir2020dynamic}, but data presents additional challenges, as cloth simulation is struggling to replicate realistic cloth dynamics and real-world data is very difficult to annotate.
% %In robotics, tasks such as \textit{grasp point detection}, \textit{unfolding}, and \textit{cloth category classification} have so far used a limited amount of simulated and/or real-world data, where the simulation of textiles and their dynamics still lack realistic deformations \textcolor{red}{[Cite XXX]}. 
% Existing data used in these works are, however, not only partially or fully restricted to public use \textcolor{red}{[Cite XXX]} but are also far away from representing a continuous manipulation task from start to finish~\textcolor{red}{[Cite XXX]}.

\begin{figure}[tb!]
		\centering
		\includegraphics[width=0.5\textwidth]{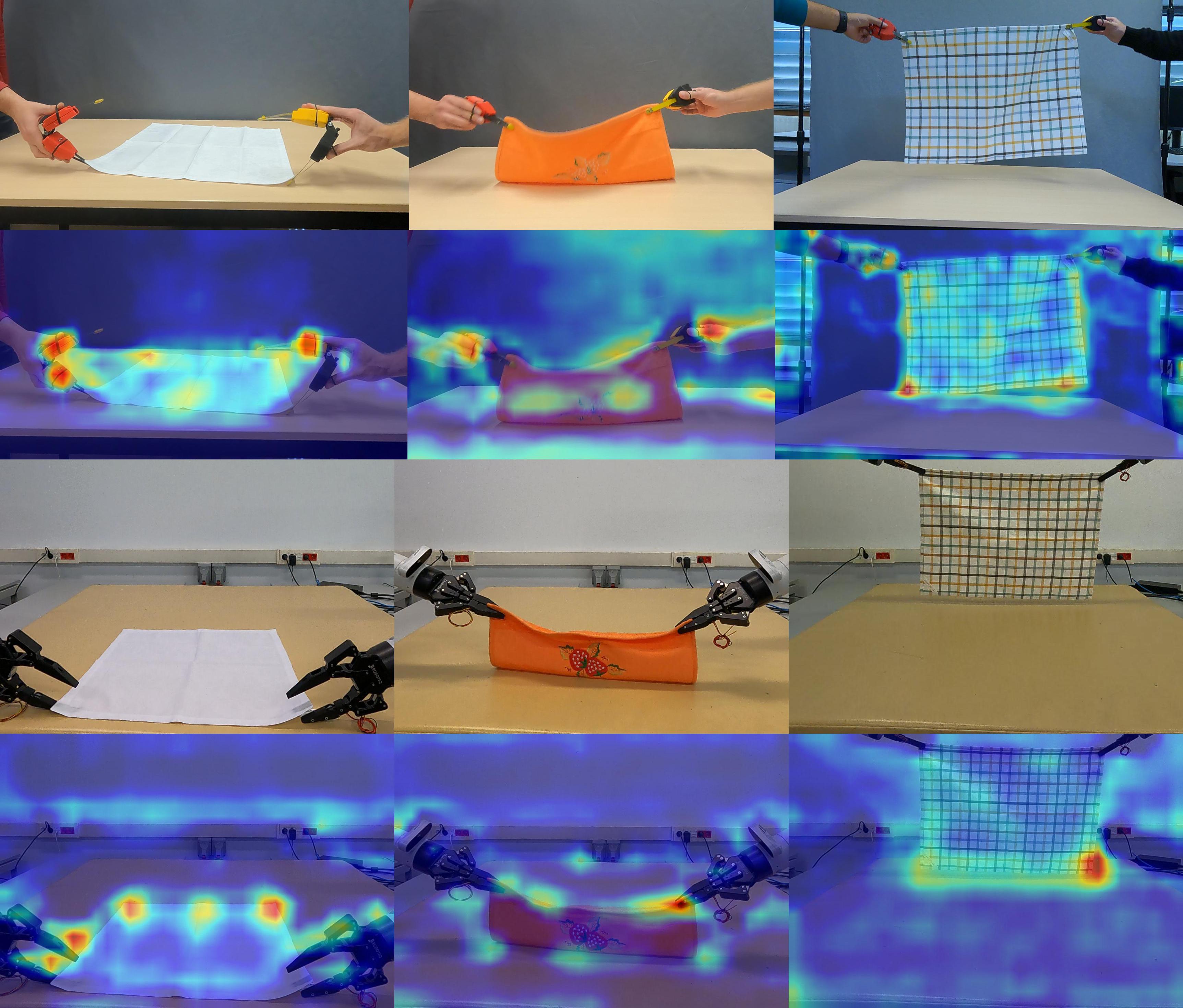}
		\caption{Correctly estimated semantic states \textit{flat}, \textit{semi-lifted with two grippers}, and \textit{lifted with two grippers} from  human and  robot   demonstrations using three unseen test cloths  \textit{White}, \textit{Orange}, and  \textit{Grid}.
		The heatmaps indicate  which regions   receive more attention, as highlighted in red. This shows that  both for the human and the robotic demonstrations	similar areas  are considered to estimate the semantic states. }
		\label{fig:overview}
\end{figure}
	
In this paper, we classify semantic cloth states of rectangular cloths during the course of a continuous manipulation task (see Fig.~\ref{fig:overview}). Cloth state estimation has been usually focused in the past to estimate the deformation state, and more in particular, the corresponding mesh \cite{li2018model, pumarola2018geometry} or interesting grasping points \cite{seita2018deep,corona2018active}. In this work, we introduce a high-level semantic description of the cloth state that includes information on not only the deformation state but also the grasping state and the contacts with the environment.
%By introducing a concept of semantic representation, we implicitly explore the deformation level on the surface of the textile.
%This gives us a unique chance to encode each perceived manipulation as a sequence of semantically different  states. 
We use the definition of the grasp type introduced in \cite{borras2020grasping}, which describes textile grasps based on the geometry of the prehension contacts, that can be either grippers or environmental objects that interact with the manipulation, such as the table where the manipulation takes place. In addition, our semantic cloth states include not only  abstract cloth deformation types (e.g. \textit{crumpled, flat, folded}), but also tags representing where the cloth is grasped from (e.g. right/left corners, edge, etc.).
%We here extend this framework to define the semantic states in continuous cloth manipulations by including labels for cloth configuration, the agent's grasping state, and environmental contacts. 
This allows us to define a sequence of semantic states during a continuous manipulation task. By being able to autonomously identify such states opens the door to learning plans of manipulations from human demonstrations, monitoring tasks, and closing the loop of a high-level planing.
%This new state definition yields a narrower re-grasping space due to the change in the semantic state of cloth configuration after having a contact with the environment.
%In a nutshell, our approach explores both the garment surface and its grasping points while estimating the semantic states.
%In our approach, the placement and pose of the gripper is of utmost importance. 
 
To investigate the scalability of our  semantic state concept, we collected a new large scale RGB image dataset captured during human demonstrations of different  \textit{uni- and bi-manual cloth manipulation} tasks such as \textit{folding diagonally} and \textit{lifting with two grippers}. 
Each captured frame is annotated by one of ten semantic states described in \autoref{tab:states}.
To solve the state estimation problem, we  employed state-of-the-art network models (e.g. EfficientNet \cite{tan2019efficientnet} and DeiT \cite{touvron2021training}) pre-trained on Imagenet \cite{deng2009imagenet}. 
Different neural networks were used for the task with the aim of providing an initial baseline on the dataset.
To show the generalization of our framework, we further recorded relatively long and complex cloth manipulation tasks   performed both by humans and by two Kinova robot arms, where we can monitor the manipulations in totally new scene contexts.

To summarize, our contributions are threefold:

\begin{itemize}
    \item We introduce a novel fully-annotated, uni- and bi-manual cloth manipulation dataset with $33.6$K RGB images involving 10 different semantic states and using $18$ different textile objects.
    \item We propose a cloth state estimation benchmark on this dataset and provide baseline experiments using state-of-the-art neural networks.
    \item We perform experimental evaluations showing that our state estimation framework can be used for monitoring robot and human demonstrations in  new scene contexts.
    %by demonstrating 4 different manipulation tasks performed by humans and/or robotic arms to monitor the states based on the predictions of a network given a manipulation plan
\end{itemize}

\section{RELATED WORK}

\subsection{Cloth Manipulation}

%In the context of cloth manipulation, scene understanding is essential for learning purposes, whether they are from human demonstrations or high level planning. Specifically for continuous manipulation of objects, the accurate prediction of the semantic states is a prerequisite for monitoring and imitation tasks. 

%Cloth manipulation is a long-standing research topic. 
%have been studied in recent years for tasks in assistive robotic tasks. 

%Julia: I remove this paragraph because it seems we need the space and assist dressing is really out of the scope of this paper
% Learning-based solutions for robotic assisted dressing have been proposed, where the focus is either on the grasping points~\cite{jimenez2020perception,zhang2020learning}  to execute the manipulation or on the probabilistic tracking and estimating the user posture~\cite{zhang2019probabilistic}. Along the same lines, collaborative learning strategies have been investigated through simulation for bi-manual assisted dressing~\cite{clegg2020learning}. In all these works, the scene is always around an articulated object, from which the deformable object will adopt it's state and discard completely the deformations prior to the dressing task \textcolor{red}{This last sentence is not clear! Rephrase!}. 

Research on the perception of textile objects and their grasping points has been done for 2D and 3D data.
For instance, by detecting task-oriented grasping points of the collar of a shirt with the use of a 3D descriptor, simple hanging tasks have been performed \cite{ramisa20163d}.
In \cite{seita2018deep}, corners of the cloth are detected in RGB-D images to perform the folding tasks.
To avoid multiple grasping strategies, active search with the use of different neural networks has been employed to recognize two grasping points \cite{corona2018active}. Finally, semantic area segmentation and domain adaptation were  used to identify grasping points from a single shot image with the help of synthetic data \cite{ren2021grasp}.
Significant progress has been made on tasks such as \textit{folding}, \textit{unfolding}, and \textit{spreading} by learning policies in a simulated environment and then transferring them into real world manipulators~\cite{puthuveetil2022bodies,seita2020deep}, or already performing all these tasks in simulation~\cite{seita2021learning}. 

In the context of image-based learning approaches, various methods have been introduced, which particularly rely on the Euclidean distance between pixels to identify equal states~\cite{hoque2020visuospatial,seita2018deep,yan2020learning,jangir2020dynamic,lippi2020latent}.  More similar to our work, the gripper states \cite{matas2018sim} or the robotic arm joints \cite{yang2016repeatable} were jointly used   with the cloth information.
These works, however, omit the interaction between the gripper and the cloth.

%Add here an extra paragraph on vision techniques using topological data analysis and deformable reconstruction  and benchmarking

%Benchmarking on highly deformable objects like garments were proposed to evaluate different grasps and manipulations \cite{garcia2020benchmarking} while 
Persistent homology was used to extract topological features of deformables with nontrivial topology in a simulated environment \cite{antonova2021sequential}. Other   works rather focused on understanding the deformation by reconstructing the scene from single shot images \cite{fuentes2018deep,pumarola2018geometry} or from the detected edges \cite{gabas2017physical}.

Along the same lines with our here presented work, abstract semantic representations of states in a manipulation task, based on contact interactions between the object, the hands, and the environment have also been introduced in the past in \cite{worgotter2013simple} using rigid objects, with the application to the manipulation recognition, segmentation \cite{aksoy2011learning} and robot execution tasks \cite{aein2019library}.

\subsection{Deformable Object datasets}

%Computer vision has been one of the fields where neural network excel and has been the dominant approach to solve end-to-end perception tasks in the past decade. An essential requirement however for most supervising learning techniques is the need of large-scale annotated datasets which improve a network's generalization capabilities. 

In the context of deformable objects and more particularly textiles, several attempts have been made to create various datasets.
Large-scale static cloth image datasets were introduced for the category classification tasks \cite{liu2016deepfashion, sun2016recognising, sun2017single}, which have also been extended by introducing landmarks to enhance the classification performance \cite{ziegler2020fashion}.

%Attempts have been made for data generation our the spectrum of manipulation tasks. 
Human demonstrations of folding tasks have been recorded  with the corresponding skeletal labels of the person performing the manipulation. These works focused on action recognition rather than state estimation \cite{verleysen2020video}. Other datasets for manipulation tasks had a limited spectrum of actions, i.e., they  omitted the gripper interactions \cite{schulman2013tracking}. Category estimation and cloth part segmentation have also been performed jointly by employing  images coupled with grasping point descriptors \cite{ramisa2012using,ramisa20163d}.

There also exist several datasets addressing the state estimation problem. Synthetic data of hanging garments~\cite{mariolis2015pose} from multi-view points generated at large scale was used for convolutional neural networks. A different approach generated a mesh of various 3D deformable objects by energy minimization from RGBD images~\cite{willimon2012energy}. Unlike ours, these datasets are, however, not available for   public use and thus limit their applications in different use-cases.

%%%%%%%%%%%%%%%%%%%%%%%%%%%%%%%%%%%% table %%%%%%%%%%%%%%%%%%%%%%%%%%%%%%%%%%%%%%%%%%%%%%%%%%%%%%%%%%%
\begin{table}[t]
\flushleft
\caption{Manipulation tasks and semantic states.}\label{tab:states}
\includegraphics[width=\linewidth]{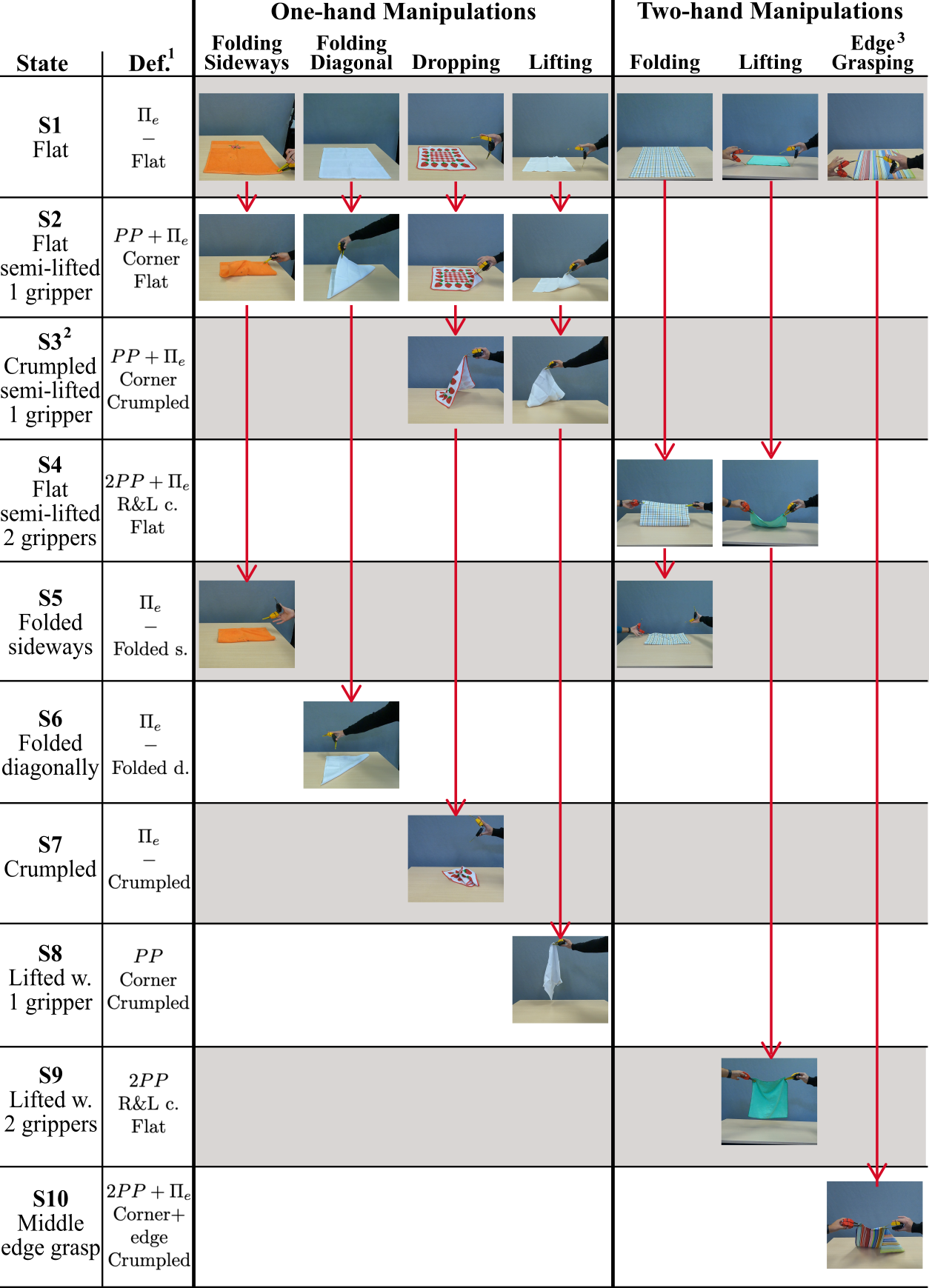}
\footnotesize{ 1: For each state, we define the grasp type, the location of the grasps and the semantic description of deformation. Following~\cite{borras2020grasping}, $PP$: pinch grasp, $2PP$: bi-manual pinch, and $\Pi_e$: the extrinsic contact with the table.\\
2: By crumpled we mean the cloth is deformed enough so that it cannot go back to a flat configuration without additional manipulation, as opposed to flat in the S2 state, that can be reversed to its previous state.\\
3: This manipulation is repeated at different distances from the corner grasp, always on the same edge.}
\end{table}
%%%%%%%%%%%%%%%%%%%%%%%%%%%%%%%%%%%%%%%%%%%%%%%%%%%%%%%%%%%%%%%%%%%%%%%%%%%%%%%%%%%%%%%%%%%%%%%%%%%%%

\section{METHOD} 

Following the work in~\cite{borras2020encoding}, we focus on human demonstrations which are mainly performed with two point grippers, each is handled by a different subject to grasp and manipulate a cloth simultaneously  (see Fig.~\ref{fig:overview}). 
The deformable object manipulated in our experiments is either a towel or a kitchen cloth with various patterns, sizes, and stiffness. The cloth size is the only restriction that is imposed, since we assume that the cloth should fit within the table edges while being completely open and flat as shown in the first row in Table~\ref{tab:states}. 
The position of the cloth is usually placed to the center of the table, unless the action is performed solely with one gripper.
In this uni-manual grasping case, the cloth is placed closer to the human demonstrator holding  the gripper  to ensure that the cloth is within the gripper's reach.

%Actions were performed through human demonstration on point grasp manipulation which is described in \cite{borras2020grasping}. Two point grippers handled by two different persons standing on the opposite sides of the table where used to manipulate the object. The actions were predefined before performing the manipulations and the grasps were two grippers require to grasps the object were performed simultaneously. The deformable of our choice was either a towel or a tablecloth of various patterns and sizes.

\subsection{Cloth Manipulation Tasks and Semantic States} 
\label{sec:manipulation}
Seven different \textit{uni- and bi-manual cloth manipulation} tasks such as \textit{folding diagonally} and \textit{lifting with two grippers}
 are performed on eighteen different cloths. Since not all cloths are square shaped, we consider the manipulations that involve diagonally folding only with cloths that are  square.
% or have their opposite edges of equal length. %%--> not sure what you mean, but that is either true for all rectangular, or is equivalent to be squared 
%
Table~\ref{tab:states} shows in each column one of the seven manipulation types.  
There are also in total ten semantic states, each of which  defines a unique deformation type of the cloth, as introduced in the grasping-centered framework in \cite{borras2020grasping,borras2020encoding}. The definition of the states takes inspiration from works like \cite{aksoy2011learning} where each change of contact interaction between hand, object and environment was considered a different scene state, however, in our case changes in deformation category are also considered.
%These semantic states are unique and define the deformation type of the manipulated cloth.    
%
Rows in Table~\ref{tab:states} depict these semantic states with the definition of the corresponding grasp type, location of grasp in the cloth, and deformation category. 

Each manipulation   is composed of a sequence of semantic states. For instance, the uni-manual cloth manipulation \textit{folding sideways} (shown in the first column in Table~\ref{tab:states} involves three states: \textit{S1: flat},  \textit{S2: flat semi-lifted with one gripper}, and  \textit{S5: folded sideways}. On the other hand, as shown in the last column in Table~\ref{tab:states}), the bi-manual manipulation \textit{edge grasping} has only two states:  \textit{S1: flat} and  \textit{S10: middle edge grasp}.
Note that for the sake of clarity, each manipulation task in  Table~\ref{tab:states}  is shown with a different textile object from our proposed dataset.

\begin{figure}[b!]
    \centering
    \includegraphics[width=\linewidth]{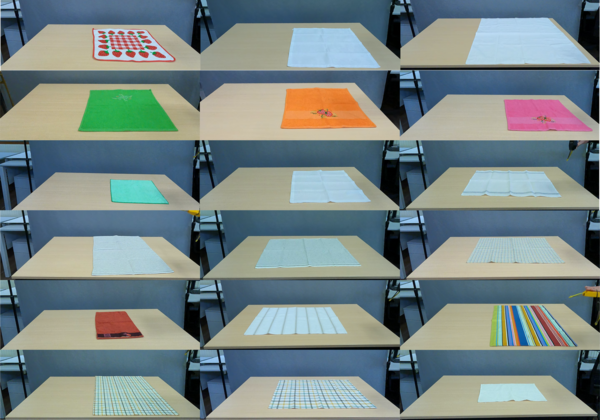}
    % \includegraphics[width=\linewidth]{figures/squared.png}
    % (a)
    % \includegraphics[width=\linewidth]{figures/rectangulars.png}
    % (b)
    \caption{Textiles used in our dataset. The first 8 are squared, and the rest rectangular.}
    \label{fig:textileObjects}
\end{figure}

\subsection{Data Collection and Annotation}
\label{sec:dataset}
We create a large scale dataset showing various human manipulation demonstrations on  different cloth types. During each human demonstration, an RGB video is recorded. Each extracted frame is then manually annotated with one of ten  semantic states described in \autoref{tab:states}.

%During each demonstration, videos are captured and then used to extract each frame. The semantic states of the manipulation are labeled manual through perception only. In that regard some states are labeled based on judgement of the person who does the labeling rather than some universal criterion. In that sense some data include noisy labeling which is impossible to be avoided. Further, in order to ease the heavy process of data annotation, for bi-manual manipulations we accept that the grasp of the grippers occurs at the same frame and thus removing a maximum 3 frame window in case grasps don't occur at the same time.

%%%%NOT COMPLETE UNTIL I KNOW THE GARMENT SIZES ALSO I'LL ADD THE FIGURE SHOWING THE GARMENTS%%%%%%

%Each of these seven manipulation tasks is performed between two and four times using   eighteen different cloths. 
At the start of each demonstration, the cloth is placed in the initial state, i.e., lying flat on the table. For the sake of having more natural scenes, the initial flat position of the cloth also contains slight deformations such as wrinkles. 
%Furthermore, several manipulations are performed on each cloth in different speed to increase the diversity  in observed deformations. 

To increase the scale of the dataset, each of the  seven manipulation tasks   is performed at least two times by altering the speed, the initial position, the grasping points, and the manipulation trajectory. In order to introduce a higher volume of deformation, eighteen different garments were introduced of different size or shape, while each has a unique color texture and pattern  as shown in Fig.~\ref{fig:textileObjects}. Seven of these garments are of squared shape, ten of rectangular,  and one is squared but with smoothed corners. The table, the grippers, and the background remain the same for all manipulations. The RGB camera position is most of the time static, with minimal changes across the manipulations. 

After having 22 human demonstration recordings using seven different manipulation types and eighteen different garments, we collect in total $33.6$K fully annotated RGB images of $10$ various semantic states. 

The annotation has been done manually through human observation. To easy the heavy workload of data labelling, the images are annotated once it exhibits a state change. Otherwise, the remaining image frames are automatically labelled with the last adjacent state name. Furthermore, some short and flickering states, which  either last less than $3$ frames or are  not in our state list given in Table~\ref{tab:states}, are omitted.

%transitions which produce semantic states outside of the 10 described in \autoref{tab:states} are not represented and are either removed if they are less than 3 frames or labelled as a neighbouring state(i.e. in the robotic demonstrations at the end of a folding manipulation, when the gripper releases the garment  from a certain height, the state is automatically labelled as \textbf{folded sideways} even though the corners are still on the air and falling). It also has to be noted that the dataset is not noise free. Boarderline cases such as the transition from the \textbf{Flat semi-lifted 1 gripper} to \textbf{Crumpled semi-lifted 1 gripper} where completely based sole on the observation through the images and thus making impossible the use of any geometric criterion when the state change occurs.

%%%%%% EXTRA INFORMATION BASED ON THE GARMENTS WILL BE ADDED HERE%%%%%% I NEED TO GO AND MEASURE THE SIZE OF THE CLOTHES WE USED%%%%%%%%%%%%%%%%%%%

%By performing the seven continuous manipulation tasks, we were able to generate and label ten semantic states.
%The semantic states are labeled by following the grasping-centered framework introduced in \cite{borras2020grasping,borras2020encoding}. 

%%%%%%% EXTRA INFORMATION BASED ON THE ROBOTIC MANIPULATION ONCE IT'S DONE%%%%%%%%%%%%%%

\subsection{Semantic State Estimation}
\label{sec:networks}
Given the annotated dataset introduced in section~\ref{sec:dataset}, we employ state of the art neural networks to estimate the semantic states of the manipulated deformable objects. For this purpose, we particularly use networks pre-trained on Imagenet \cite{deng2009imagenet} with the help of transfer learning. 
%
%To verify our approach we used neural networks for our state estimation problem. 
%To ensure that the approach can be easily replicated without the need of vast resources, we particularly use  neural networks pre-trained on Imagenet \cite{deng2009imagenet} with the use of transfer learning.
%\textcolor{red}{Not understandable what you mean in this next statement:}
% To ensure that our comparisons are meaningful without bias, we also opted for neural networks who use have the same input resolution. 
% Note that the classifier layer is the same across all networks. 

The networks of our choice are relying on convolutional operations such as  EfficientNet \cite{tan2019efficientnet}, ResNet-50 \cite{he2016deep} and ResNeXt-50 \cite{xie2017aggregated}. % instead of the current popular vision transformers ViT \cite{dosovitskiy2020image}.
%\textcolor{red}{Not understandable what you mean in this next statement:} However, to ensure that we get results by a different method of extraction, we use the more lightweight DeiT \cite{touvron2021training}.
We also use a more lightweight vision transformer-based model DeiT \cite{touvron2021training}.
Note that we append the same classifier layer  to these network models and employ the cross-entropy loss function.  In addition, we augment the data by randomly translating, flipping around the y-axis, and cropping. We also apply random rotation  with a 15 degree restriction. 
The optimizer of our choice was stochastic gradient with warm restarts \cite{loshchilov2016sgdr}.

\begin{table}[!b]
\centering
\caption{Quantitative Evaluation. \label{tbl:acc}}
\begin{tabular}{l|c|cccc}
         & \multicolumn{1}{c|}{Validation} & \multicolumn{4}{c}{Test Scores}  \\  
Networks & \multicolumn{1}{c|}{Scores} & \multicolumn{1}{c}{White} & \multicolumn{1}{c}{Orange}  & \multicolumn{1}{c}{Grid} & Average \\ \hline
ResNet-50~\cite{he2016deep}            &  97.77          & \textbf{96.27}  &  91.79          &   95.54 & 94.53  \\ %\cline{1-1}
ResNeXt-50~\cite{xie2017aggregated}    &  97.50          & 95.67           &  92.30          &   95.51  & 94.49\\ %\cline{1-1}
EfficientNet~\cite{tan2019efficientnet}&  97.71          & 95.53           &  \textbf{93.16} &   \textbf{96.61}       & \textbf{95.10}\\ %\cline{1-1}
DieT~\cite{touvron2021training}        &  \textbf{98.20} & 93.85           &  86.18          &  91.60          & 90.54 \\ \hline
\end{tabular}
\label{tab:acc}
%\vspace{-4mm}
\end{table}
%%%%%%%%%%%%%%%%%%%%%%%%%%%%%%%%%%%%%%%%%%%%%%%%%%%%%%%%%%%%%%%%%%%%%%%%%%%%%%%%%%%%%%%%%%%%%%%%%%%%%%%%%%%%%%%%%%%%%%%%%%%%%%%%

\section{EXPERIMENTS}
We evaluate the performance of semantic state estimation networks in two use cases: monitoring of human  and robot manipulations. In both cases, we measure the correct classification rate (i.e., the accuracy scores) as the evaluation metric. Furthermore, we generate class activation maps  of the networks to ensure that the  network attention is on similar features in both human demonstrations and robot executions. % (Fig.~\ref{fig:overview}). 

\subsection{Human Demonstrations for Training}
\label{sec:human_training}
We trained the four networks in section~\ref{sec:networks} by dividing our  $33.6$K annotated dataset with a 75-25 stratified split. 
To verify that the trained networks are not biased towards any specific cloth type, we exclude all the manipulations with some  specific cloth types from the training data and employ them only for  testing purposes. For instance, all demonstrations with the cloth  \textit{White}  shown in Fig.~\ref{fig:overview}  are reserved as the new unseen test data, while all the remaining data are used for training. The same leave one out testing protocol is also used for the  \textit{Orange} and \textit{Grid} garments shown in Fig.~\ref{fig:overview}.
%To ensure however that the network is not overdepended on the garment, we remove before training all the manipulations of a cloth which will be used as a test set. 

Table~\ref{tab:acc} shows accuracy scores  in percentage (\%) for the validation and individual test cases with these three cloths.  
As  shown in Table~\ref{tab:acc}, EfficientNet~\cite{tan2019efficientnet} performs the  best on average in contrast to the other three networks. Having a minor difference between the validation and average test scores confirms that EfficientNet is not biased with the cloth types in the training data. 
The sample activation maps provided in Fig.~\ref{fig:overview} depict  regions (such as corners, edges, etc.) where EfficientNet pays more attention when predicting the correct states, in these three  test cases. %Note that grasp points and cloth edges have received particular attentions.   
Note that the network tends to focus more on the grasp points, cloth edges and corners (reddish zones) rather than the cloth textures and patterns. This is a strong evidence indicating that the state estimates heavily rely on manipulation-related features such as grasp points and cloth edges instead of irrelevant cues such as texture.

\begin{figure}[!t]
	\centering
	\includegraphics[width=0.5\textwidth]{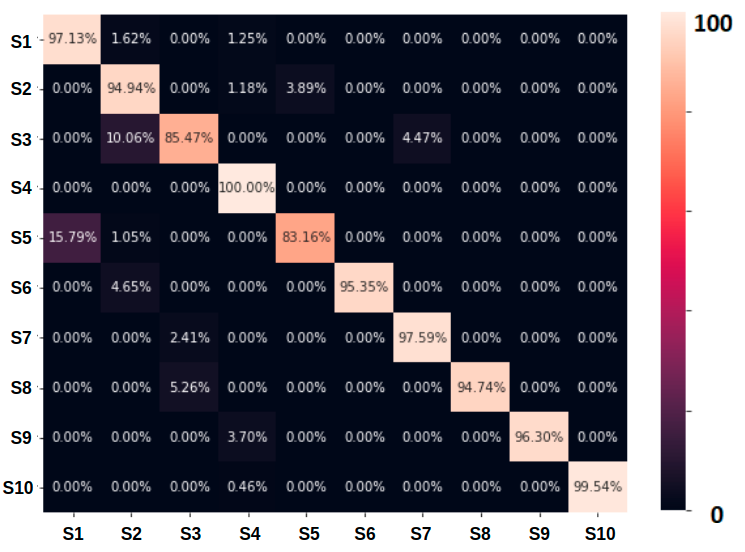}
	\caption{Confusion matrix of EfficientNet  on the unseen test cloth  \textit{White}  in  Table~\ref{tab:acc}. 
	Each raw corresponds to the one semantic state in Table~\ref{tab:states}. For instance, $S1$ represents \textit{Flat} and \textit{S2} refers to  \textit{Flat semi-lifted with one gripper}.}
	\label{fig:confusion matrix}
\end{figure}

Fig.~\ref{fig:confusion matrix} depicts the confusion matrix for the test case of the \textit{White} cloth in  Table~\ref{tab:acc}. This figure
clearly shows that there is no major confusion between the predicted semantic states even when the network is exposed to a new test cloth.

\begin{figure}[b!]
		\centering
        \includegraphics[width=0.48\textwidth]{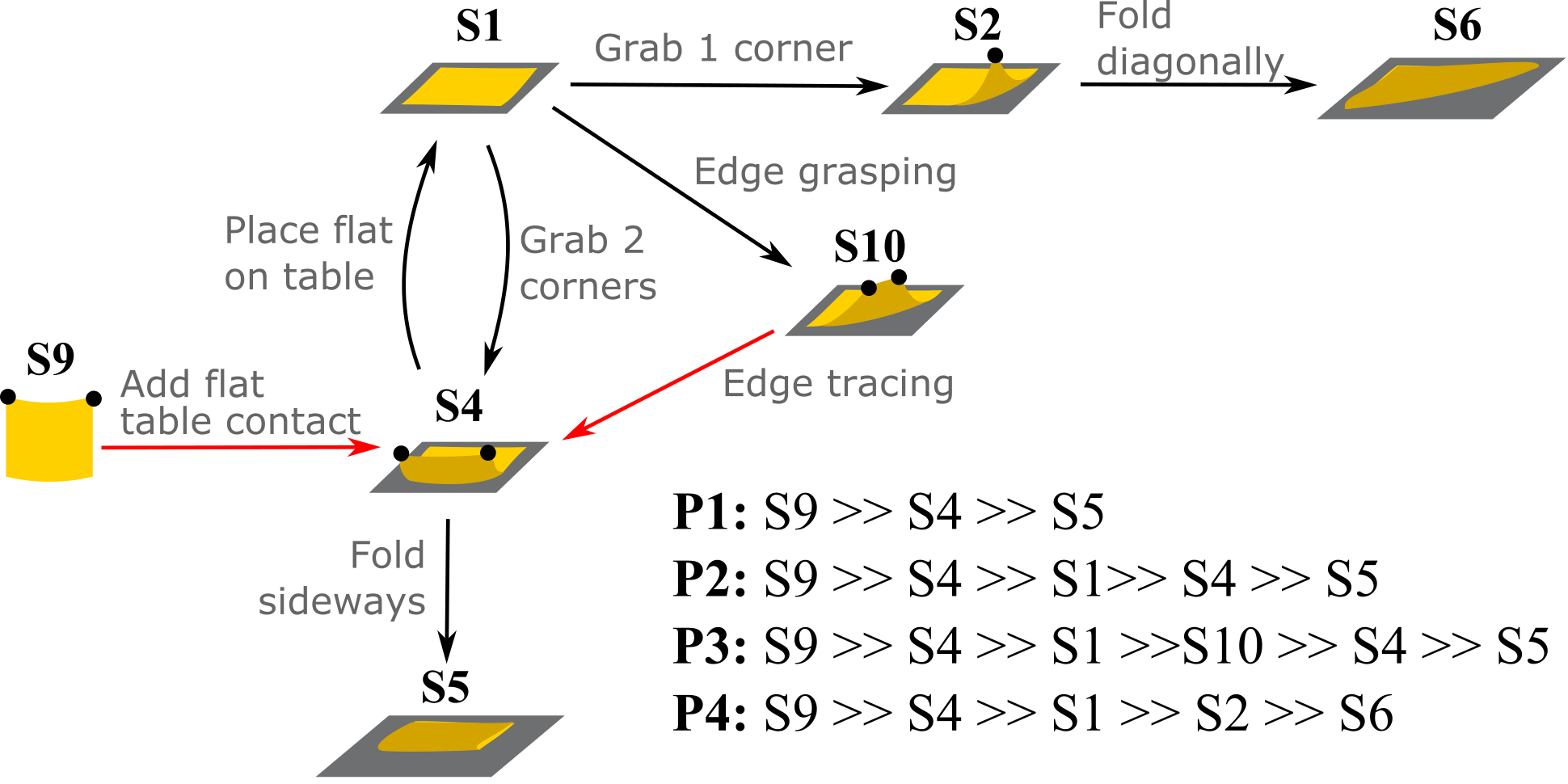}
		\caption{The four long manipulation scenarios presented as a graph of the semantic states. Following the state definitions in Table~\ref{tab:states}, in these four test demonstrations, the goal states are either \textit{S5: Folded sideways} or \textit{S6: Folded diagonally}.
		% Initially, all test scenarios begin with the same state \textit{S9: Lifted with two grippers}. This state is followed by \textit{S4: Flat semi-lifted with two grippers}. Next, there are different paths, each continues with one of the goal states.} 
		All possible paths to reach the goals are shown in the bottom right corner of the figure. Red arrows correspond to state transitions that are not present in our proposed dataset.}
		\label{fig:graph}
\end{figure}

\begin{figure*}[t!]
    \centering
    \includegraphics[scale=0.5,trim={2cm 0 2cm 5cm},]{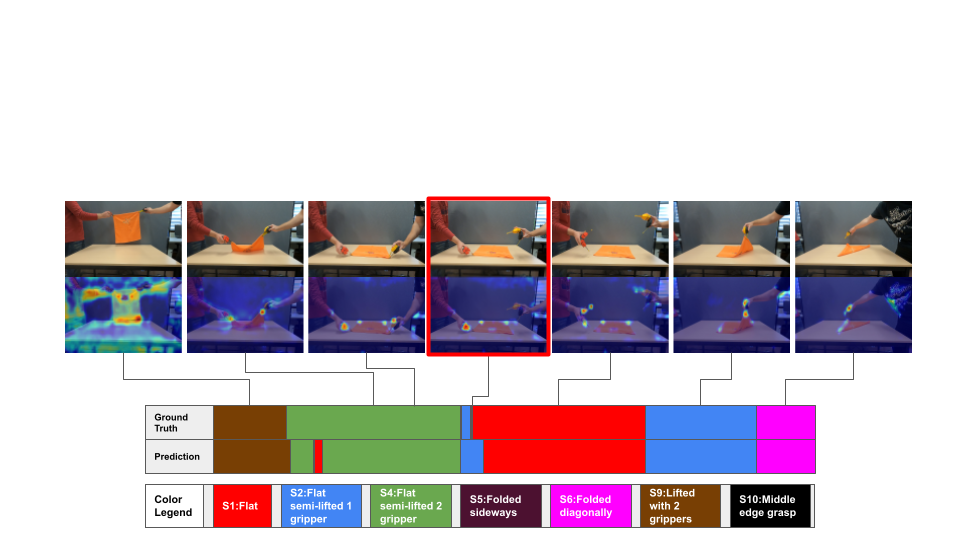} 
    \caption{Human demonstration of the manipulation scenario P4   described in Fig.~\ref{fig:graph}. The colored blocks represent the network predictions vs   human labeled ground truth. Sample images with the corresponding class activation heatmaps are depicted on the top.  The image in the red frame represents the borderline case which is  incorrectly classified as S2 whereas the ground truth is S1.  
    %The images above and their corresponding class activation heatmap are linked to their prediction class.
    %From those 7 images, six are predicted correctly while the 4th image from left to right is incorrectly classified as S1:Flat while the ground truth is S2:Flat semi-lifted with 1 gripper.}
    }
    \label{fig:humandemo}
\end{figure*}

\subsection{Long and Complex Human Demonstrations for Testing}
\label{sec:human_test}
To show the generalization of our state estimation network, we further recorded relatively long and complex cloth manipulation scenarios performed   by humans. Our ultimate aim here is to diagnose the capacity of our network in totally new scene contexts. 

For this purpose, we have recorded four chained manipulation scenarios %with the same three cloths, 
each is composed of different number of manipulation tasks and semantic states defined in Table~\ref{tab:states}. We used the same three cloths (\textit{White}, \textit{Orange}, and \textit{Grid}) in these test scenarios. Fig.~\ref{fig:graph} illustrates those four scenarios as a graph sequence. Here, graph nodes represent the semantic states, whereas edges describe the manipulation tasks to go from one state to the next. For instance, the first scenario represents \textit{folding a cloth sideways} by following a path (P1 in Fig.~\ref{fig:graph}) with three states: \textit{S9,~S4} and \textit{S5}, whereas the third scenario follows P3 with six states to reach the same goal: \textit{S9,~S4,~S1,~S10,~S4} and \textit{S5}. Note that all  scenarios start with  the same state \textit{S9: Lifted with two grippers} to reach two different goal states, either \textit{S5: Folded sideways} or \textit{S6: Folded diagonally}.

In these four novel scenarios, we obtained $52.16\%$ average accuracy score for all semantic states. The reason of this substantial drop in accuracy is mainly due to having totally new deformation and manipulation types, such as \textit{Add flat table contact} and \textit{Edge tracing} (shown in red arrows in Fig.~\ref{fig:graph}), neither of which can be encapsulated  by one of our trained semantic states or manipulation tasks. However, after fine-tuning the EfficientNet model with \textit{only two} additional human demonstrations of these new deformation types using garments from the training set, the overall accuracy increased up to $84.80\%$. 
We here note that during these four  scenarios, we collected more than 6K image frames, and only $350$ RGB images from a garment that is not part of the three testing garments were used for the fine-tuning operation.
%tzelepis: The training set is the same that I used before. The same weights which were used to produce table 2 were applied in all task. Afterwards for the increased accuracy I added around 200 frames of edge tracing and 150 frames of lowering. Now for the test the orange is 2.436, the white 2.028 and the grid which doesn't include the P4 task 1647

This experimental finding clearly shows the complexity of the state estimation problem: having unseen cloth deformations can still lead to errors due to high dimensionality in the deformation  of  textile objects. Incrementally refining the already trained network can, however, boost the performance to a large extent, as explored in our experiments.

Fig.~\ref{fig:humandemo} illustrates the EfficientNet network performance on the manipulation scenario P4  described in Fig.~\ref{fig:graph}. The colored blocks clearly show that the network predictions are very similar to the human defined ground truth. There still exist false positive predictions which particularly emerge around borderline cases, as highlighted by the red frame in Fig.~\ref{fig:humandemo}. For instance, the network has difficulty to distinguish states \textit{S1:Flat} and  \textit{S2:Flat semi-lifted with 1 gripper}, in particular, while the subject is about to release the garment.    
The provided sample activation maps in Fig.~\ref{fig:humandemo} also depict that the network pays more attention to the grasp points, cloth edges and corners as expected.

\begin{figure*}[t!]
    \centering
    \includegraphics[scale=0.6,trim={5cm 0 5cm 0},]{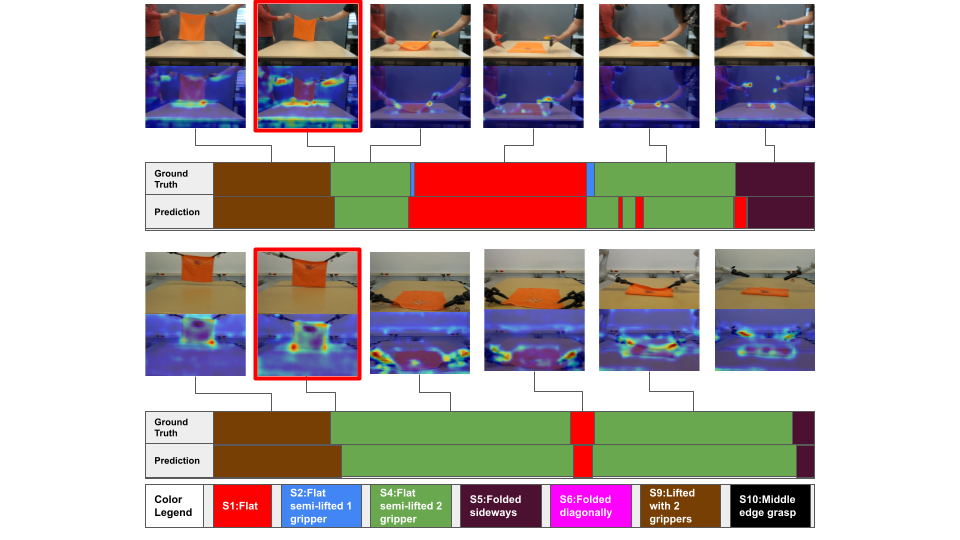} 
    \caption{Human and robot demonstrations of the manipulation scenario P2 described in Fig.~\ref{fig:graph}. The colored blocks in both cases represent the network predictions vs   human labeled ground truth. Sample images with the corresponding class activation heatmaps are depicted on the top.  The images in the red frames represent false positive predictions. }
    \label{fig:intervals}
\end{figure*}

\subsection{Robot Manipulations}
We executed the first two  scenarios (P1 and P2) of these four long test scenarios described in Fig.~\ref{fig:graph} using 2 Kinova robot arms. In both scenarios, the goal is to fold the garment sideways by initially holding it in the air with both grippers.

We recorded the robot executions of P1 and P2 using 5 garments,   3 of them being the same test  garments: \textit{White}, \textit{Orange}, and \textit{Grid}.
We collected in total 24K frames from robot executions. For the test garments, we had about 14.5K frames.
For performance measurement of our network, we follow  the same leave one out testing protocol employed in section~\ref{sec:human_training}. In each test case, we excluded data including one of \textit{White}, \textit{Orange}, and \textit{Grid} garments respectively and used all the remaining recordings for fine-tuning our EfficientNet model already trained with the human demonstration data in section~\ref{sec:human_test}. 

Note that differences in scene contexts between the robot and human demonstrations introduce the domain shift problem. Therefore, our EfficientNet trained with the human demonstration data in section~\ref{sec:human_test} cannot be directly employed here to monitor these robot executions. To easily overcome this problem, we applied this additional fine-tuning operation before testing with three  garments. Note also that  the recorded images were first cropped to have similar views with those recorded in human demonstrations. Finally, we obtained $97.52\%$  average accuracy on the unseen three test garments: \textit{White}, \textit{Orange}, and \textit{Grid}. The reason of obtaining such as a high accuracy is also due to the fact that robot motions are slow and less noisy than that of humans.

Fig.~\ref{fig:intervals} shows the network performance for human and robotic demonstrations of the same manipulation scenario  P2  described in Fig.~\ref{fig:graph}. Each image is accompanied by the activation map taken from the network's final convolution layer, which helps us  observe the similarities between both demonstrations. It is evident that the network focuses on the lower boundary of the garment while being held in the air. Once the garment has contact with the table, the attention shifts to the grippers' position with respect to the corners and the edges of the garment.
The red frames depict incorrect predictions, which emerged, for instance, when  the garment just switched from states \textit{S9:lifted with two grippers} to \textit{S4:flat semi-lifted with two grippers}. 
The colored blocks in Fig.~\ref{fig:intervals} clearly show that such false positive predictions are borderline cases where the state is either about to change or just switched to the next.

 \begin{figure}[b!]
    \centering
    \includegraphics[width=0.48\textwidth]{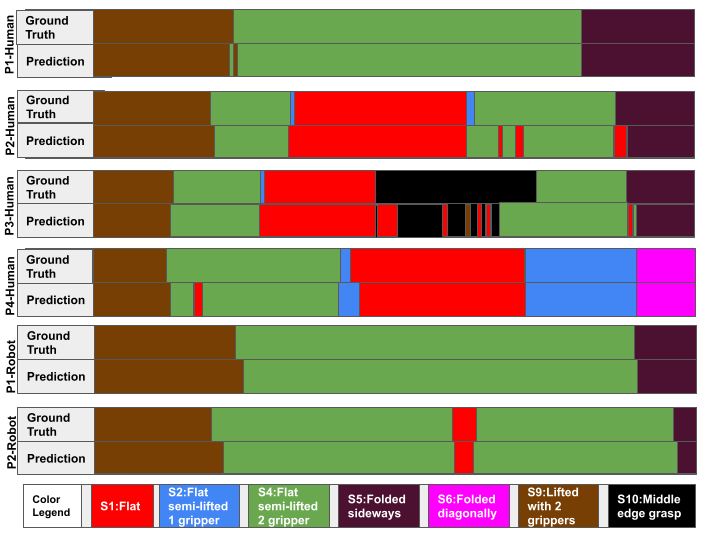}
    \caption{ Network predictions vs   human labeled ground truth for all long manipulation scenarios  described in Fig.~\ref{fig:graph}. The first four are from human demonstrations and the last two are from robot executions.  }
    \label{fig:bars}
\end{figure}
 
Fig.~\ref{fig:bars} displays the network performance  together with the human
labeled ground truth for all human and robot demonstrated long manipulation scenarios described in Fig.~\ref{fig:graph}.
This side-by-side comparison indicates that the network can successfully predict semantic states even in the case of having unseen textiles. Note that the lengths of manipulation scenarios are normalized for the sake of clarity in the display.
This figure also shows that the state predictions in robot executions is less noisy than that of human demonstrations, since humans follow more natural motion patterns. Note that the false network predictions in other paths (P3-Human and P4-Human in Fig.~\ref{fig:bars}) also emerge mostly around state transitions. 

\addtolength{\textheight}{-4.5cm}
\section{DISCUSSION}
 
We first would like to highlight the fact that in this work, we do not propose any novel network model. We rather show that the state-of-the-art models (e.g., EfficientNet \cite{tan2019efficientnet}) can already handle the challenging state estimation problem with the help of transfer learning. Our reported high accuracy scores in Table~\ref{tab:acc} already show that there is no need to focus on designing new deep network architectures. Instead, we should  address the domain shift problem. 
Our results show how the network loses accuracy very fast when presented with new unseen deformations, due to the high dimensionality of the cloth state estimation problem. We, however, show that fine-tuning the network with very few new frames can again boost the performance.
Despite the domain shift, the class activation heatmaps in Figs.~\ref{fig:humandemo} and ~\ref{fig:intervals} also  clearly ensure that our network  focuses on similar manipulation relevant regions such as the grasp points, cloth edges and corners instead of the cloth textures and patterns.

In our dataset, human demonstrations were performed by two different individuals who had grippers attached to their hands and  executed the planned actions through vocal commands. This process introduced additional noisy states  to the expected state transitions in Fig.~\ref{fig:graph}. For instance, a noisy state \textit{S2} appeared in Figs.~\ref{fig:humandemo} since one person released the gripper faster than the other subject. These naturally emerging noisy states in human demonstrations are kept in our dataset to make the dataset more challenging.   

It is evident from Figs.~\ref{fig:humandemo},~\ref{fig:intervals}, and ~\ref{fig:bars}  that the network's wrong predictions are mostly due to borderline cases  where two consecutive states are very similar to each other. We here note that false state estimates, which are irrelevant to borderline cases (e.g., see P3-Human in Fig.~\ref{fig:bars}), can easily be solved by incorporating temporal state information during the monitoring task. Since our networks perform frame-wise predictions, the temporal cue is omitted in this work.     

Furthermore, we pose the following questions to better understand our network performance:
 
\textbf{What if we exclude the fine-tuning process?} In this case, we observe substantial accuracy drops both in human and robot test manipulations. Our ablation study shows that the fine-tuning boosts the performance by $32.64\%$ and $45.06\%$  for human and robot demonstrations, respectively.  

\textbf{What if we train only with the robot demonstrations?} In case of excluding the human demonstrations and fine-tuning the network only using robot demonstration also leads to an increase in the average accuracy by $1.08 \%$. However, the generated activation class heatmaps become extremely noisy. Therefore, we can conclude that this slight increase in accuracy comes with the cost of  over-fitting to the observed scene. Instead of collecting more robot executions, learning the semantic states from more natural human demonstrations is needed to regularize the network and solve the overfitting problem.  

\textbf{What if there exists no fine-tuning data?} We argue that such cases can be handled by injecting the uncertainty estimation to the prediction. However, such approaches are going beyond the scope of this paper.

%\item  Also, propose some questions like what if we do not apply fine tuning or what if we test with human demos right after fine tunining with robot demos...what if we also introduce novel manipulations in the test case...

%Further the results raised several questions, like what if we do not apply fine-tuning for states that are the same but the cloth deformation is very different or what if we test it with human right after we fine tune with the robots demonstrations. Further answers are required on how can the network's uncertainty be handled in case we introduce new manipulations and states in the test case and how can this affect the monitoring process.

\section{CONCLUSIONS}

In this paper, we presented and evaluated a novel framework to solve the problem of semantic state estimation in continuous cloth manipulation tasks in an end-to-end manner. Our semantic state definition differs from classic rigid object approaches in that we introduce a high-level semantic description of the cloth state which couples the cloth  deformation type,  the
grasping state and the contacts with the environment. To validate our approach, we benchmarked four different networks on our new  dataset of 33.6K annotated RGB images.

As a future work, we plan to enlarge our dataset  by introducing a higher variety of deformable shapes (semantic states) and more complex manipulation tasks. Furthermore, we would like to incorporate the depth cue which can capture  geometrical information in the scene, and thus, play a crucial role to autonomously define  unseen textile deformations and plan a proper grasping accordingly.    

We hope that the here presented dataset and benchmarks  will be adopted by the cloth manipulation community and  trigger further contributions in robotics. 


%%%%%%%%%%%%%%%%%%%%%%%%%%%%%%%%%%%%%%%%%%%%%%%%%%%%%%%%%%%%%%%%%%%%%%%%%%%%%%%%

%\section{REMAINING TASKS}
%\textcolor{red}{ 
%\begin{itemize}
%    % \item Reference \cite{ramisa2014learning} has missing author names.
%    \item To save space, go through the reference list and shorten some of those which has redundant abbreviations and %info. For instance, \cite{deng2009imagenet} can be shortened by removing ieee 2009 which appear twice.
%    \item Be consistent: cloth/cloths/clothes/garment...which one is correct!
%\end{itemize}
%}

%%%%%%%%%%%%%%%%%%%%%%%%%%%%%%%%%%%%%%%%%%%%%%%%%%%%%%%%%%%%%%%%%%%%%%%%%%%%%%%%

%%%%%%%%%%%%%%%%%%%%%%%%%%%%%%%%%%%%%%%%%%%%%%%%%%%%%%%%%%%%%%%%%%%%%%%%%%%%%%%%
%\section*{APPENDIX}


%\section*{ACKNOWLEDGMENT}

%%%%%%%%%%%%%%%%%%%%%%%%%%%%%%%%%%%%%%%%%%%%%%%%%%%%%%%%%%%%%%%%%%%%%%%%%%%%%%%%

%References are important to the reader; therefore, each citation must be complete and correct. If at all possible, references should be commonly available publications.

\bibliographystyle{IEEEtran}
\bibliography{projectbib}

\end{document}